\documentclass{bmvc2k}
\usepackage{float}
\usepackage{soul}
\usepackage{xr}
\makeatletter
\newcommand*{\addFileDependency}[1]{
  \typeout{(#1)}
  \@addtofilelist{#1}
  \IfFileExists{#1}{}{\typeout{No file #1.}}
}
\makeatother

\newcommand*{\myexternaldocument}[1]{
    \externaldocument{#1}
    \addFileDependency{#1.tex}
    \addFileDependency{#1.aux}
}

\myexternaldocument{supplementarymaterial}
\title{Weakly Supervised Training of Monocular 3D Object Detectors Using Wide Baseline Multi-view Traffic Camera Data}

\addauthor{Matthew Howe}{matthew.howe@adelaide.edu.au}{12}
\addauthor{Ian Reid}{ian.reid@adelaide.edu.au}{1}
\addauthor{Jamie Mackenzie}{jamie.mackenzie@adelaide.edu.au}{2}

\addinstitution{
 Australian Institute for \\Machine Learning\\
 University of Adelaide\\
 Adelaide, Australia
}
\addinstitution{
 Centre for Automotive Safety Research \\
 University of Adelaide\\
 Adelaide, Australia
}

\runninghead{Howe, Reid, Mackenzie}{WIBAM}

\def\etal{\emph{et al}\bmvaOneDot}

\begin{document}

\maketitle

\begin{abstract}
Accurate 7DoF prediction of vehicles at an intersection is an important task for assessing potential conflicts between road users. In principle, this could be achieved by a single camera system that is capable of detecting the pose of each vehicle but this would require a large, accurately labelled dataset from which to train the detector. Although large vehicle pose datasets exist (ostensibly developed for autonomous vehicles), we find training on these datasets inadequate. These datasets contain images from a ground level viewpoint, whereas an ideal view for intersection observation would be elevated higher above the road surface. 
We develop an alternative approach using a weakly supervised method of fine tuning 3D object detectors for traffic observation cameras; showing in the process that large existing autonomous vehicle datasets can be leveraged for pre-training. 
To fine-tune the monocular 3D object detector, our method utilises multiple 2D detections from overlapping, wide-baseline views and a loss that encodes the subjacent geometric consistency.
Our method achieves vehicle 7DoF pose prediction accuracy on our dataset comparable to the top performing monocular 3D object detectors on autonomous vehicle datasets. We present our training methodology, multi-view reprojection loss, and dataset.
\end{abstract}

\section{Introduction}

% \hl{[RF: Matt -- there are no citations on your introduction]}

The majority of collisions between road users in urban areas occur at intersections. Visual surveillance could have a significant impact on providing tools to understand conflicts between road users. This work presents a model to infer location, size, and yaw rotation (7DoF pose) of vehicles from monocular images. This will enable road safety researchers to build 3D models of traffic flow and understand the severity, likelihood, and frequency of conflicts to enable better design and flow-control of intersections.

Current methods of training models to collect road users' 7DoF pose rely on expensive sensors such as LiDAR and large amounts of hand labelled data. For these reasons, existing methods are not cost effective when performing multi-intersection studies; a common method for intersection safety evaluations. It is becoming increasingly frequent for road safety researchers to utilise monocular cameras to collect road user behaviour data \cite{vid-cycleinfra, vid-highwayramp, vid-intervru, vid-movesafe, vid-pedvehicleint}. Since 3D object detectors for road safety are not readily available methods utilise ground plane calibration and off the shelf 2D object detectors to approximately localise vehicles passing through intersections. The ability to easily collect precise location, rotation, and space occupancy data of road users is invaluable for accurate measurements of safety and will open the door to a broader application of 3D vision in road safety.

To utilise the simplicity and low-cost of monocular video collection, an object detector must be able to infer the 7DoF pose of vehicles from a frame.
Typically, a monocular 3D object detector (mono3DOD) for vehicles is trained on autonomous vehicle data which contains thousands of hand annotated 3D objects that utilise LiDAR to accurately label 7DoF poses. 
There is a domain gap between ego vehicle data and intersection observation cameras. This is due to the perspective difference between cameras mounted at vehicle roof height versus elevated above the road surface.
While datasets from traffic observation viewpoints exist, they do not contain LiDAR data or 3D object annotations that would be needed to train a mono3DODs conventionally.

In this paper we develop a method which is able to utilise inexpensive sensors and requires no additional manual labelling to train a mono3DOD. We leverage an existing (pre-trained) mono3DOD designed for ground-level detection, and show how this can be fine-tuned in a weakly supervised manner using multi-view camera data. Using our methodology, road safety researchers can fine-tune a model for their camera setup through the temporary installation of additional overlapping cameras for 10 minutes and achieve 7DoF pose prediction accuracy that is comparable to fully supervised models.

\textbf{Summary of contributions}
\begin{enumerate}
    \item We develop a multi-view reprojection loss to train 3D object detectors in a weakly supervised fashion to an accuracy comparable to top performing mono3DODs.
    \item We provide the wide baseline multi-view (WIBAM) dataset. Captured using monocular cameras surrounding an intersection, with automated annotations and a hand labelled test set.
\end{enumerate}

\section{Related Work}
In this section we discuss mono3DODs and how the common datasets used to train them limit their application in road safety. We then review research how multi-view geometric constraints can be used in lieu of ground truth labels, namely in 3D pose estimation and depth prediction.

\textbf{Monocular 3D Object Detectors}: Monocular 3D object detection aims to infer the location of objects in the scene from a single image. For vehicle 3D object detectors, roll and pitch are set to zero with a flat ground assumption. Mono3DODs commonly predict an object's location in the image frame, yaw rotation, size, and depth. Differing methods utilise key points and CAD models \cite{DeepMANTA, RTM3D, MonoGRNet, MonoGRNetV2, SMOKE}, others utilise a tight fit constraint of a between a 2D and 3D bounding box \cite{3DDeepBox, multiviewreprojectionarch, Shift-RCNN, FQNet, GS3D}, and some make direct predictions of 7DOF pose \cite{Mono3D, disentanglemono3dod, SS3D, M3D-RPN, D4LCN, centernet, centretrack}. These monocular 3D object detection methods are trained with full supervision on autonomous vehicle datasets which are expensive to collect and hand label.

\textbf{Autonomous vehicle datasets}: Training of mono3DODs commonly use the autonomous vehicle datasets NuScenes \cite{nuscenes2019}, KITTI \cite{KITTI2012CVPR}, Waymo \cite{waymo_open_dataset}, and Lyft \cite{lyft2019}. These datasets were collected using cameras on top of an ego vehicle's roof. The angle between a line from the camera centre to the centre of another vehicle with reference to the ground plane, reffered to as elevation angle, for autonomous vehicles is around 5 degrees. Traffic cameras observe vehicles at elevation angles as large as 50\textdegree. Mono3DODs have not been exposed to vehicles observed at these elevation angles during training. This results in noisy and inaccurate predictions of 7DoF pose. To train mono3DODs for high elevation angles in the conventional manner road safety researchers would need access to LiDAR data and extensive hand labelling which is often unobtainable. 

\textbf{Multi-view datasets}: Existing publicly available multi-view traffic datasets with overlapping field of view (FoV) cameras either have too little data to train a mono3DOD \cite{multi-viewmulti-class2011, Ko-PER} or images from elevations that were too low \cite{Carfusion, Occlusion-Net}. 

Beyond traffic based datasets, there have been several multi-view pedestrian datasets that contain overlapping FoVs, such as WILDTRACK and Toulouse campus dataset \cite{WILDTRACK, Toulouse}. However, these datasets do not contain 3D annotations or traffic scenes that would be required for our use case.

\textbf{Geometric weak-supervision of monocular object detectors}: In the context of object detection, geometric multi-view constraints were used to train pose and depth estimators without ground truth data. Simon \etal \cite{handkeypointreco} annotate 3D hand poses using 2D hand pose detector and 3D reconstruction. A monocular hand key point detector is then trained on these annotations. Predictions of pose detectors can be compared over multiple views to calculate a consistency loss that can train a monocular model in a weakly supervised manner \cite{Monocular3Dhumanposemultiview, weaklysup-multiview3Dpose}. Additionally, a multi-view geometric constraint can be used as a supervision signal for training depth prediction models \cite{geomtorescue}. Reprojection consistency losses have also been used for 3D shape prediction of vehicles in the case of Occlusion-Net \cite{Occlusion-Net}. Occlusion-Net utilises a multi-view reprojection loss where predicted object-centric 3D shapes are reprojected onto other views and compared with 2D key point detections to train models in a weakly supervised manner.

These weakly-supervised methods all aim to incorporate the additional information from multiple views to improve the predictions of object-centric pose but do not attempt to exploit the multi-view data to train a mono3DOD for 7DoF pose. We introduce a method to utilise weak geometric supervision to train a mono3DOD.

\section{WIBAM dataset}\label{sec:wibamdataset}
The wide baseline multi-view (WIBAM) dataset collected for the purposes of training our weakly supervised model contains sequential image data from multiple cameras of an inner-city four-leg intersection. This intersection has dedicated right turn lanes, large divides between counter flowing traffic, and a steady flow of traffic. Cyclists, pedestrians, cars, and utility vehicles all use this intersection as a corridor through the city. A satellite image of the intersection and samples of the dataset can be found in Section \ref{supsec:WIBAMsamples} of the supplementary material.

Four GoPro Hero 7 cameras were set to use "linear field of view mode" (i.e. GoPro's radial distortion correction), and mounted at roughly the same height on the traffic poles in the four corners of the intersection. We maximise the inter-camera FoV overlap by pointing them towards the middle of the intersection. This allows constant observation of vehicles from multiple views. Unsynchronised video data was recorded at $50$FPS by each camera at a resolution of $2560\times1440$ pixels, for roughly 15 minutes over the same period, in daylight conditions on a clear day.

Traffic light signals, which were observable from all the camera views, were used to synchronise the recordings. Synchronisation was done semi-automatically with knowledge of the locations of the traffic lights in each view. When a change in traffic light is observed in a leading video the frames in the trailing videos are dropped until they have observed the light change. Figure \ref{fig:light_synchronisation} in the supplementary material shows a graph of the before and after delay between videos across the dataset. The maximum delay was reduced from 16 frames ($0.3$s) down to $3$ frames ($0.06$s). 
% Despite synchronising the cameras at the light changes, the synchronisation may still differ in the time between light changes ($60$ seconds). However, we found later on that further data cleanup was not necessary.

After synchronisation videos are down sampled to $12.5$ FPS at a resolution of $1920 \times 1080$ resulting in $8,273$ images per camera; $33,092$ images in total. The images were split into three subsets for training, validation, and testing each containing 75\%, 15\%, and 10\% of the dataset, respectively. Each subset is separate in time such that the test and validation set contain none of the same vehicles. We hand annotate $12\%$ of the test split, as outlined in Section \ref{sec:labellingtool}. Intrinsic and extrinsic camera calibration as well as a homography matrix for the ground plane are included with the dataset. These were found using measured correlated points between the camera planes and ground plane. We use the perspective-n-point algorithm to solve extrinsic calibration and the linear least means squares to solve the homography matrix.

\subsection{Automatic multi-view annotations}\label{sec:autoannotations}
We automatically annotate vehicles using an off the shelf 2D object detector \cite{Detectron2018}. In order to train a mono3DOD with multi-view geometric supervision, associations of a vehicle's 2D bounding boxes between the views must be known. Clustering is used to associate detections corresponding to the same vehicle across cameras. Specifically, rays are cast through each camera's 2D detections to find where the ray intersections the ground plane. These points of intersections are then assigned to vehicles via DP-means clustering \cite{DP-means}. Examples of the output of the clustering algorithm can be found in Figure \ref{fig:clustering examples} of the supplementary material.

\subsection{Automatically annotated dataset information}
The WIBAM dataset contains automatic annotations for vehicles detected in all images. A total of $45,762$ automatically annotated multi-view vehicles, leading to $116,702$ 2D bounding boxes of vehicles are contained in the dataset. There are $31,333$ vehicles viewed from two cameras, $3,680$ from three, and $10,749$ from four. Note that the high number of two camera observations are due to many predictions occurring as vehicles are approaching and leaving the intersection. When vehicles are within the intersection we observe that, in the majority of cases, all four cameras can observe the vehicle. In the case where a vehicle is occluded, by a bus for example, results in three camera detections. We estimate that each vehicle is viewed an average of $45$ times resulting in a total of $1,000$ unique vehicles captured at different times, viewing angles, and locations throughout the intersection. 

\subsection{Hand labelled test set}\label{sec:labellingtool}
In order to quantify the performance of our weakly supervised method we developed a test set using our multi-view annotation tool. The tool is used to annotate vehicle 7DoF pose and visibility by viewing the reprojections of annotations onto each camera view. A screenshot of the labelling tool GUI is included in Section \ref{supsec:labellerGUI} of the supplementary material.

The WIBAM test set which contains $400$ hand annotated images resulting in $1,651$ hand labelled 7DoF vehicle annotations, of which $400$ are partly occluded. It is important to note that when vehicles are visible in fewer than three views it is challenging to accurately manually label them. However, in the more frequent case where three or four cameras observe the vehicle manual annotations can be produced using our tool. Distributions of detected vehicles' distances, elevation angles, and visibilities be found in Section \ref{supsec:testset} of the supplementary material.

\section{Multi-view loss}\label{sec:multiviewloss}
\begin{equation}
\label{eqn:multiview}
L_{MV} = \frac{1}{n}\sum_{n=1}^{n} (1 - GIoU_{n}) + L_{focal}
\end{equation}
The multi-view reprojection loss \ref{eqn:multiview} is compromised of two functions which are summed together. A focal loss for measuring the discrepancy between the predicted vehicle centre points and the centre points predicted by the 2D object detector, as in CenterNet \cite{centernet}. This component is responsible for fine tuning the predicted 2D object centre detections from the mono3DOD so that it learns to identify vehicles in the new domain.
The second function is our geometric consistency loss which utilises 3D object detections' reprojections onto other cameras to calculate the Generalised Intersection over Union (GIoU) \cite{GIOU}; encouraging consistency of the 3D predictions between views.

\begin{figure}
    \centering
    \includegraphics[width=\textwidth]{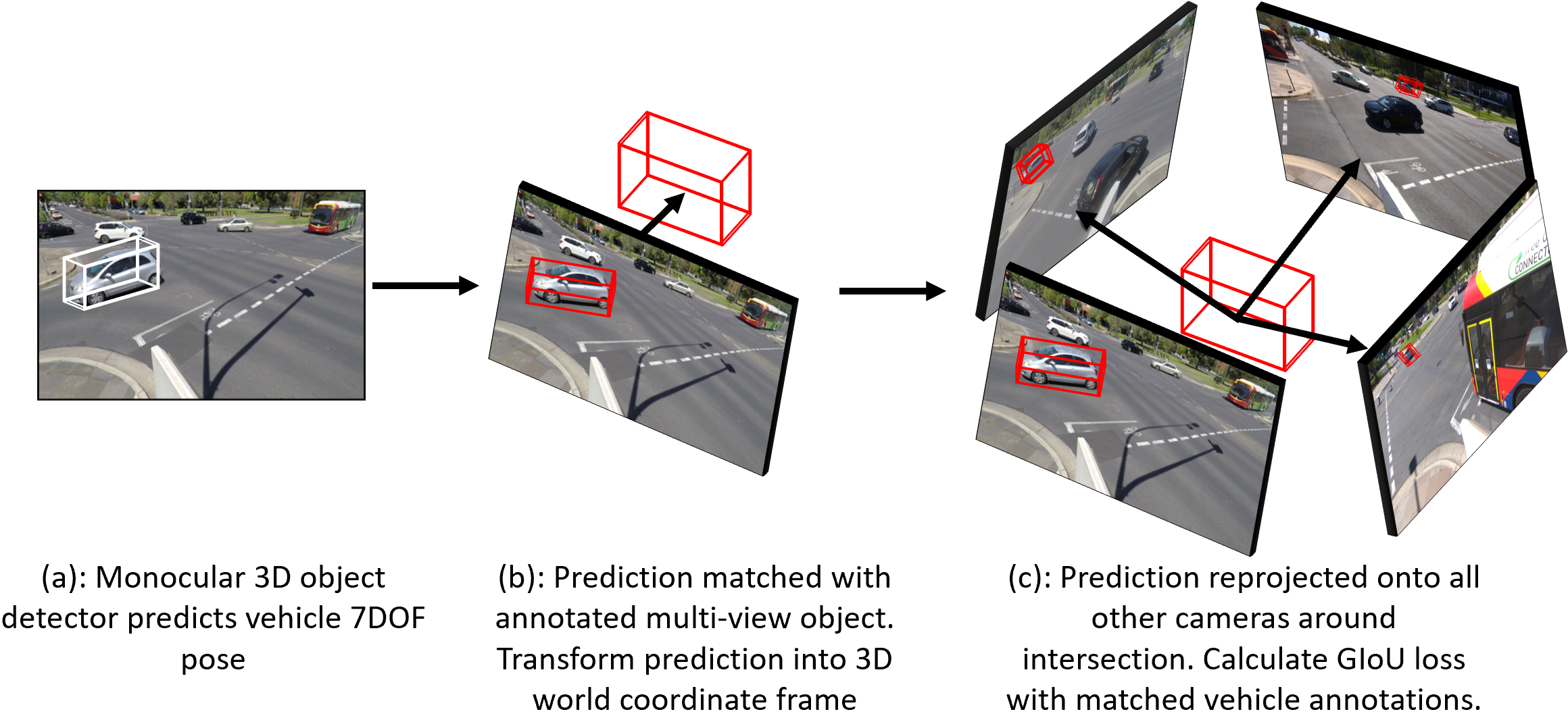}
    \caption{Method for calculating the reprojection loss at train time}
    \label{fig:traintime_diagram}
\end{figure}

Our multi-view loss is calculated using the monoc3DOD to make predictions on one of the images from a single time frame, see Figure \ref{fig:traintime_diagram}(a). Each vehicle prediction is then matched with the automatic multi-view annotations described in Section \ref{sec:autoannotations}. Using this match, the 3D object is transformed from a camera-centric to a world coordinate frame, see Figure \ref{fig:traintime_diagram}(b). The vehicle in world coordinate frame is then projected onto all other camera views of the scene as shown in Figure \ref{fig:traintime_diagram}(c). A tight fitting 2D bounding box around the reprojection can then be compared with automatic multi-view 2D object detection boxes using GIoU.

\begin{figure}
    \centering
    \resizebox{\textwidth}{!}{
    \begin{tabular}{cc}
        \bmvaHangBox{\includegraphics[width=0.5\textwidth]{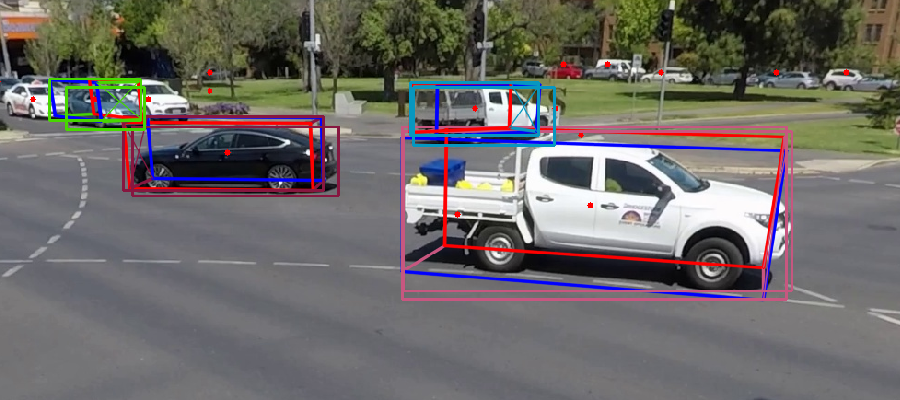}}&
        \bmvaHangBox{\includegraphics[width=0.5\textwidth]{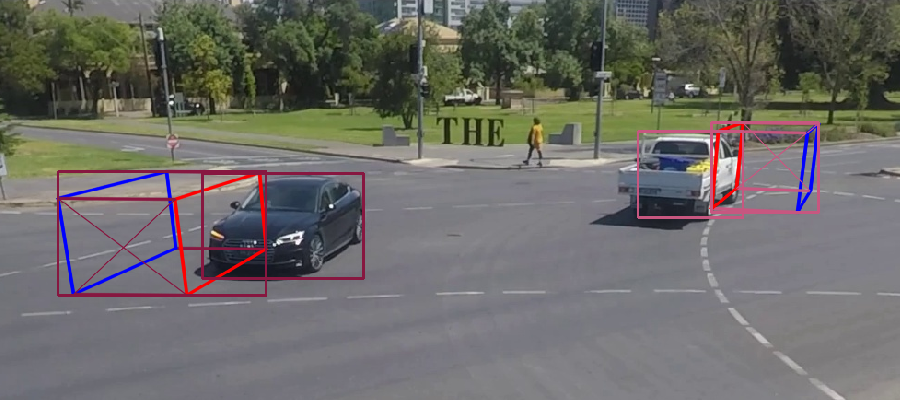}}\\
        (a)&(b)
    \end{tabular}
    }
    \caption{Training images with 3D bounding box reprojections (a) 3D detections reprojected on the input image (b) 3D detections reprojected onto the second camera}
    \label{fig:twoute}
\end{figure}

Figure \ref{fig:twoute} highlights that simply training with a single camera is insufficient to improve 7DoF pose predictions. Explicitly, a prediction using the image from (a) as input results in what appears to be a good prediction. However, once this pose is projected onto a camera from a separate viewpoint, shown by (b), we visually inspect a large error.

\section{Fine-tuning a 3D object detector}
We use a DLA-34 \cite{DLA} model from CenterNet \cite{centretrack} pre-trained on NuScenes \cite{nuscenes2019} as our mono3DOD 'baseline model'. We fine tune the baseline model on the WIBAM dataset with a batch size of 80 images on two Nvidia V100 GPUs with an initial learning rate of $3.125e^{-5}$. We drop the learning rate by a factor of 10 when validation loss plateaus for 4 epochs with a tolerance of $\pm0.001$. The fine-tuned model is referred to as the 'WIBAM model'.

\subsection{Quantitative results}
We evaluate the performance of the mono3DOD by comparing the predictions to the manually labelled test set data; see Section \ref{sec:labellingtool}. We consider errors in the different degrees of freedom separately: average translation error (ATE), average scale error (ASE), and average orientation error (AOE). Translation error is measured using the euclidean distance between the centres of the predicted and ground truth cuboids. Scale error is reported as $1-IoU$ of the ground plane bounding boxes after aligning the rotation and translation. This is equivalent to the absolute proportional difference in ground plane area. Orientation error is quantified using the smallest yaw angle between the ground truth and predicted orientations. Additionally, We calculate the 3D IoU and birds eye view (BEV) IoU to give a holistic measurement of 7DoF predictions.
% Additionally, a common error for predictions are cases where vehicles are detected to have rotations which are 180 degrees out. We refer to them as aliased rotations and consider any vehicle rotations which have an error larger than $\pm170$ degrees to be aliased.

\begin{table}[]
    \centering
    \resizebox{\textwidth}{!}{
        \begin{tabular}{|c|c|c|c|c|c|}
            \hline
            Method & 3D IoU $\uparrow$ & BEV IoU $\uparrow$ & ATE (m) $\downarrow$ & ASE $\downarrow$ & AOE (deg) $\downarrow$ \\
            \hline
            Baseline & 0.25 (s=0.18) & 0.33 (s=0.25) & 1.51 (s=0.97) & \textbf{0.12 (s=0.06)} & 5.67 (s=5.47)\\
            \hline
            WIBAM model & \textbf{0.48 (s=0.09)} & \textbf{0.64 (s=0.13)} & \textbf{0.43 (s=0.43)} & 0.23 (s=0.07) & \textbf{4.69 (s=2.91)}\\
            \hline
        \end{tabular}
    }
    \caption{Localisation performance comparison between baseline and the fine-tuned WIBAM model}
    \label{tab:localisation}
\end{table}

\begin{figure}
    \centering
    \includegraphics[width=\textwidth]{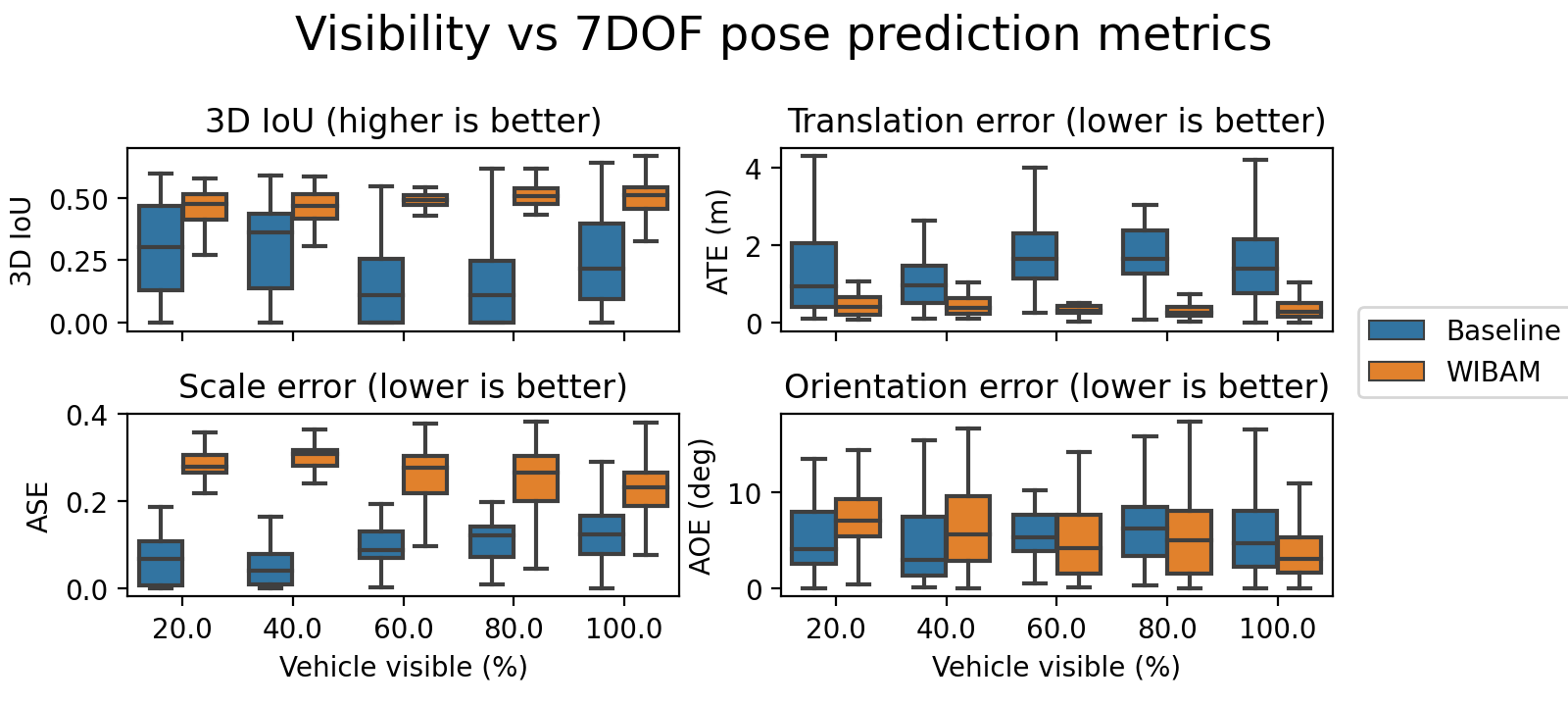}
    \caption{Effect on 7DoF pose prediction with varying amounts of vehicle occluded/truncated to camera.}
    \label{fig:visibility}
\end{figure}

Table \ref{tab:localisation} reports the performance of the baseline and WIBAM models on the WIBAM test set. Our experiments show that the 3D and BEV IoU are higher by 92\% and 94\%, respectively. This can be attributed to the lowering of ATE by 71\% and AOE by 17\%. Predictions produced by the WIBAM model induce less variance in error and IoU metrics when compared to the baseline. This demonstrates that our method of training produces predictions which are more consistent. We perform Z-tests to ensure our performance is significantly better and found that all errors in Table \ref{tab:localisation}, except scale, are significant to the 0.01 level.

We observe a higher scale error in the WIBAM model's predictions, which is believed to be caused by partially occluded vehicles. The 2D object detector used will predict a bounding box encapsulating the visible part of the vehicle. However, the full 3D bounding box predicted will be reprojected for calculation of the GIoU. This creates a signal, via the GIoU loss, encouraging the model to underestimate the size of vehicles.

We evaluated the performance of the mono3DODs when objects are occluded and truncated. Figure \ref{fig:visibility} shows how performance metrics vary as a function of vehicle visibility. Note that 3D IoU is not significantly impacted by visibility, and is consistently higher for the WIBAM model versus baseline. This can be attributed to a large reduction in translation error; see the 7DoF prediction component errors in Figure \ref{fig:visibility}.

The effect of occlusion on scale error is particularly apparent in Figure \ref{fig:visibility} where less visible vehicles have higher scale errors; a relationship not observed in baseline model predictions. We can handle these situations for truncated vehicles by clamping 3D bounding box reprojections at the image boundaries but it is not as trivial in the case of occlusions.

\begin{figure}
    \centering
    \includegraphics[width=\textwidth]{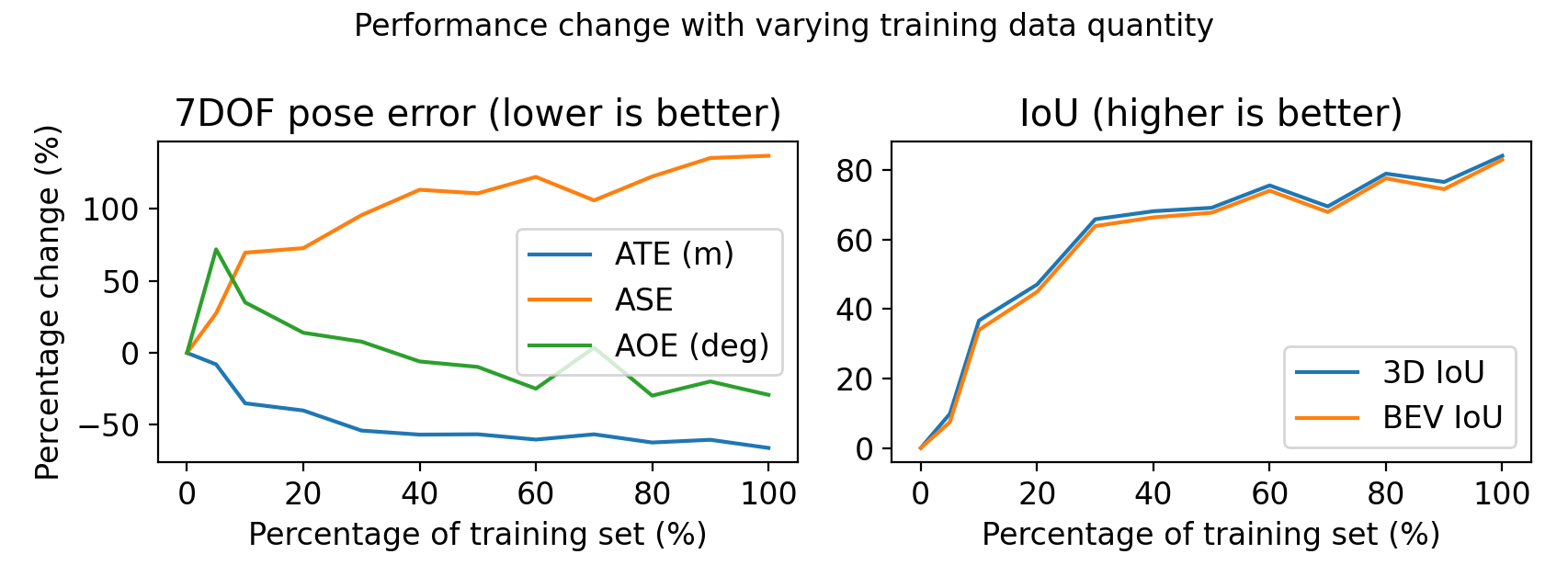}
    \caption{Training with varying levels of data. Data is shown as a percentage improvement over the baseline trained model; i.e. 0\% training set. (a) 7DoF pose components (b) Intersection over Union.}
    \label{fig:multi-train}
\end{figure}

We trained models using varying amounts of the training set to evaluate what a suitable amount of training data would be. The data selected is from the beginning of the video and is treated sequentially to simulate model fine-tuning on a given sample collected while cameras are installed. Different amounts of time, from  half a minute (5\%) up to eight minutes (100\%), are tested to see how much footage is required for our method to be effective

Training with 40\% (3 minutes 18 seconds) of the data produces the largest performance improvement on the test set; see Figure \ref{fig:multi-train}. Large orientation errors are observe when training on small amounts of data. This indicates that insufficient training data leads to poor generalisation on the test set. ATE is observed to immediately improve with any amount of data but has diminishing returns beyond 40\% of training data. A model trained with any amount of multi-view data is able to improve the predicted 3D and BEV IoU. This finding is helpful to show that for a particular distribution in time of day and vehicle flow, around 10 minutes of HD video collected at 12.5FPS is capable of returning significantly better 3D IoU.

Table \ref{tab:nuscenes-comp} shows absolute 7DoF pose errors of the WIBAM model on the WIBAM test set. We also show the current state of the art for supervised mono3DODs trained and tested on NuScences. Note that the error for each is comparable. Although this comparison is not strictly "apples to apples", we see that our method is able to use weak supervision to improve the performance so that it is comparable to the best fully supervised models.

\begin{table}
    \centering
    
    % \resizebox{\textwidth}{!}{
    \begin{tabular}{|c|c|c|c|}
        \hline
        Method & 2D ATE (m) $\downarrow$ & ASE $\downarrow$ &  AOE (deg) $\downarrow$ \\
        \hline
        WIBAM model on WIBAM test set & 0.41 & 0.22 & 4.69 \\
        \hline
        CenterNet \cite{centernet} on NuScenes cars test set & 0.47 & 0.14 & 5.38 \\
        \hline
        FCOS3D \cite{FCOS3D} on NuScenes cars test set & 0.56 & 0.15 & 5.15 \\
        \hline
    \end{tabular}
    \caption{Comparison of our 7DoF pose errors with NuScenes SOTA models}
    \label{tab:nuscenes-comp}
    % }
\end{table}

\subsection{Qualitative results}
\begin{figure}
    \centering
    \resizebox{\textwidth}{!}{
        \begin{tabular}{c}
        \includegraphics[width=\textwidth]{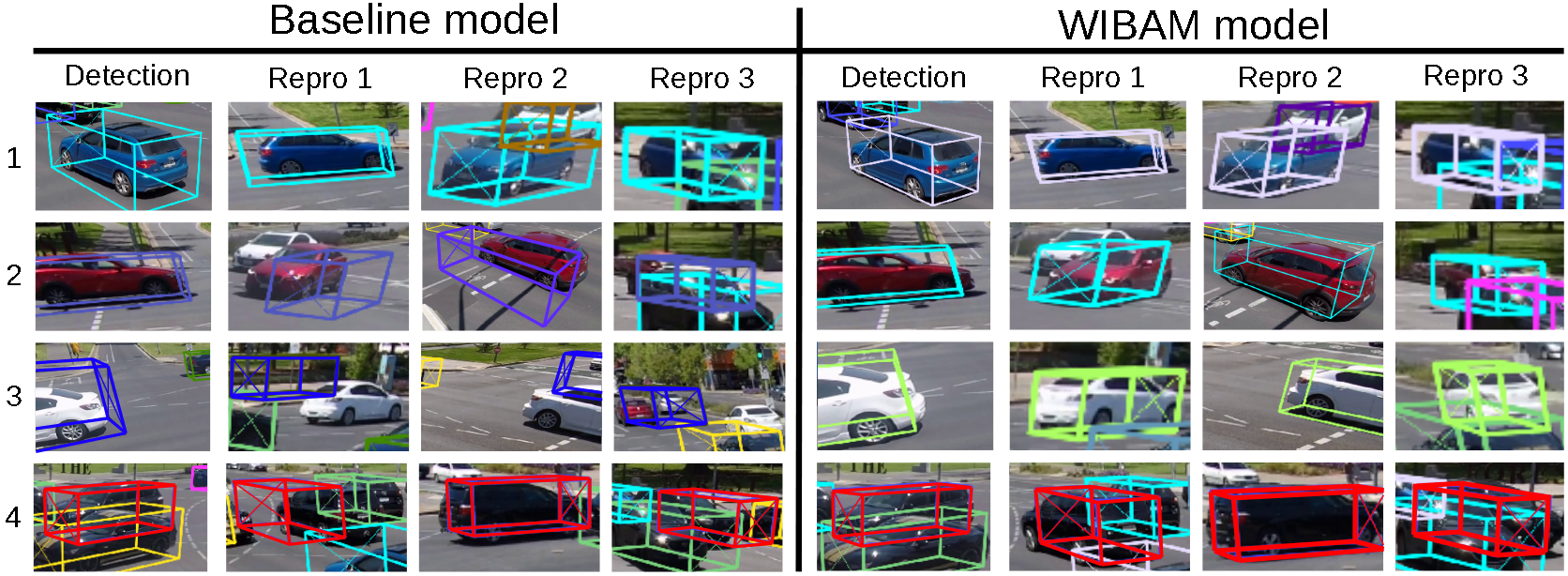}
        \end{tabular}
    }
    \caption{Detections of the same vehicles with the baseline model and the WIBAM model}
    \label{fig:qualitative-results}
\end{figure}

\begin{figure}[!h]
    \centering
    \includegraphics[width=\textwidth]{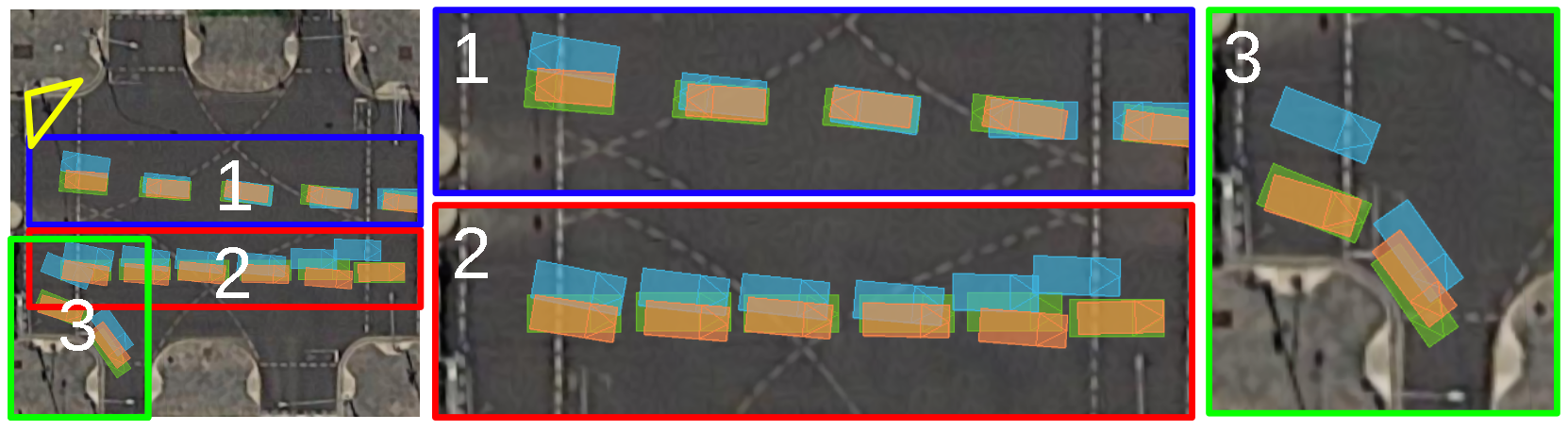}
    \caption{Multiple time instances of predicted vehicle locations in the intersection all projected onto a BEV image. Baseline predictions are shown in blue, WIBAM model predictions shown in orange, and hand annotated ground truth shown in green. Vehicles reprojections are separated, labelled, and colour coded for clarity of comparison. The yellow triangle shows the camera location. Imagery \copyright2021 Aerometrex Pty Ltd, Map data \copyright2021 Google. We provide videos of these sequences in the supplementary material.}
    \label{fig:bev_comparison}
\end{figure} 

To demonstrate our improvement in 7DoF pose accuracy, we show before and after visualisations of the baseline and WIBAM model predictions. Figure \ref{fig:qualitative-results} contains a comparison between the 3D bounding boxes produced by the baseline model and WIBAM model. Each row is a specific vehicle cropped from different viewpoints at the same time. Detection camera predictions are shown in the first columns. Subsequent columns are crops of these predictions reprojected onto the other three camera views. Row one is an example of our method's ability to refine size predictions, row two shows an example of improvements to localisation, and rows three and four contain examples of predictions on truncated and occluded vehicles.

Reprojections of 3D bounding boxes onto a camera frame often do not display the subtle errors present in model predictions. Figure \ref{fig:bev_comparison} is a BEV illustration of WIBAM and baseline model predictions projected onto the intersection. Initially estimations made for vehicle one are consistent with ground truth but a jump is observed at the end of the trajectory. Vehicle two shows that the consistency of predictions is improved with our fine tuning. Vehicle three shows a drastic improvement in localisation of a vehicle which is far away and was partially occluded by vehicle two.

Figure \ref{fig:qualitative-results} and \ref{fig:bev_comparison} illustrate the improved vehicle 7DoF pose predictions of the WIBAM model over the baseline. Our method is able to make 7DoF pose predictions of vehicles in the scene more accurately than the baseline and do so more consistently over time.

\section{Conclusion}
In this paper we have developed a weak supervision method for fine-tuning monocular 3D object detection models on high mounted, static intersection observation cameras. Specifically, our method allows a road safety researcher to fine-tune a model for a specific camera set up with 10 minutes of additional video data from cameras with overlapping fields of view and no hand labelling. Unlike long term multi-view data collection, short term multi-view data collection can be easily setup once with inexpensive, consumer grade cameras.

Our experiments show that 3D object detection models trained on ego vehicle data suffer in performance when used on intersection observation cameras. Our loss uses weak labels from a pretrained 2D object detector to ensure consistency between views of a model's reprojected 3D object detections. We show that our method of fine-tuning a monocular 3D object detector achieves vehicle 7DoF pose prediction accuracy on the WIBAM test set comparable to SOTA models on the NuScenes test set.

We also release the WIBAM dataset with 45k automatically annotated multi-view vehicles (116k 2D bounding boxes), 1,651 7DoF pose annotations of vehicles labelled with our multi-view annotation tool, and the WIBAM model. We would like to extend our method for vulnerable road users such as cyclists and pedestrians.

% For future work it would be advantageous to collect additional wide baseline multi-view training data at different intersections and varying heights with the aim to train a generic 3D object detector for static intersection observation cameras.

\section*{Acknowledgements}
This research has been supported through the Australian Government Research Training Program Scholarship. High performance compute resources used in this work were funded by the Australian Research Council via LE190100080.
% \section*{References}
\bibliography{references.bib}
\end{document}

\section{Supplementary material}
\subsection{WIBAM dataset samples}\label{supsec:WIBAMsamples}
\begin{figure}[!h]
    \centering
    \resizebox{\textwidth}{!}{
        \begin{tabular}{cc}
        \bmvaHangBox{\fbox{\includegraphics[width=.5\textwidth]{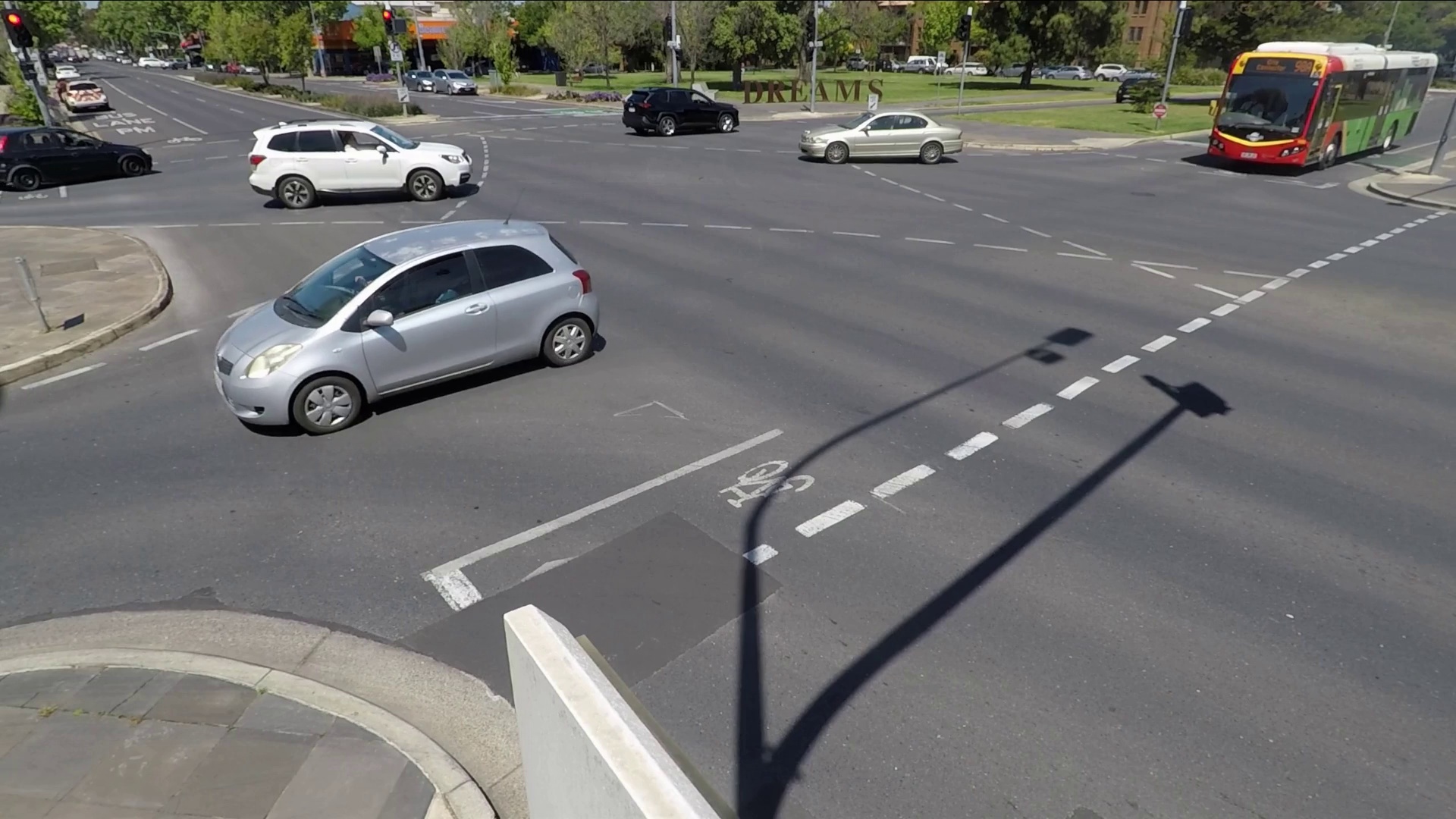}}} &
        \bmvaHangBox{\fbox{\includegraphics[width=.5\textwidth]{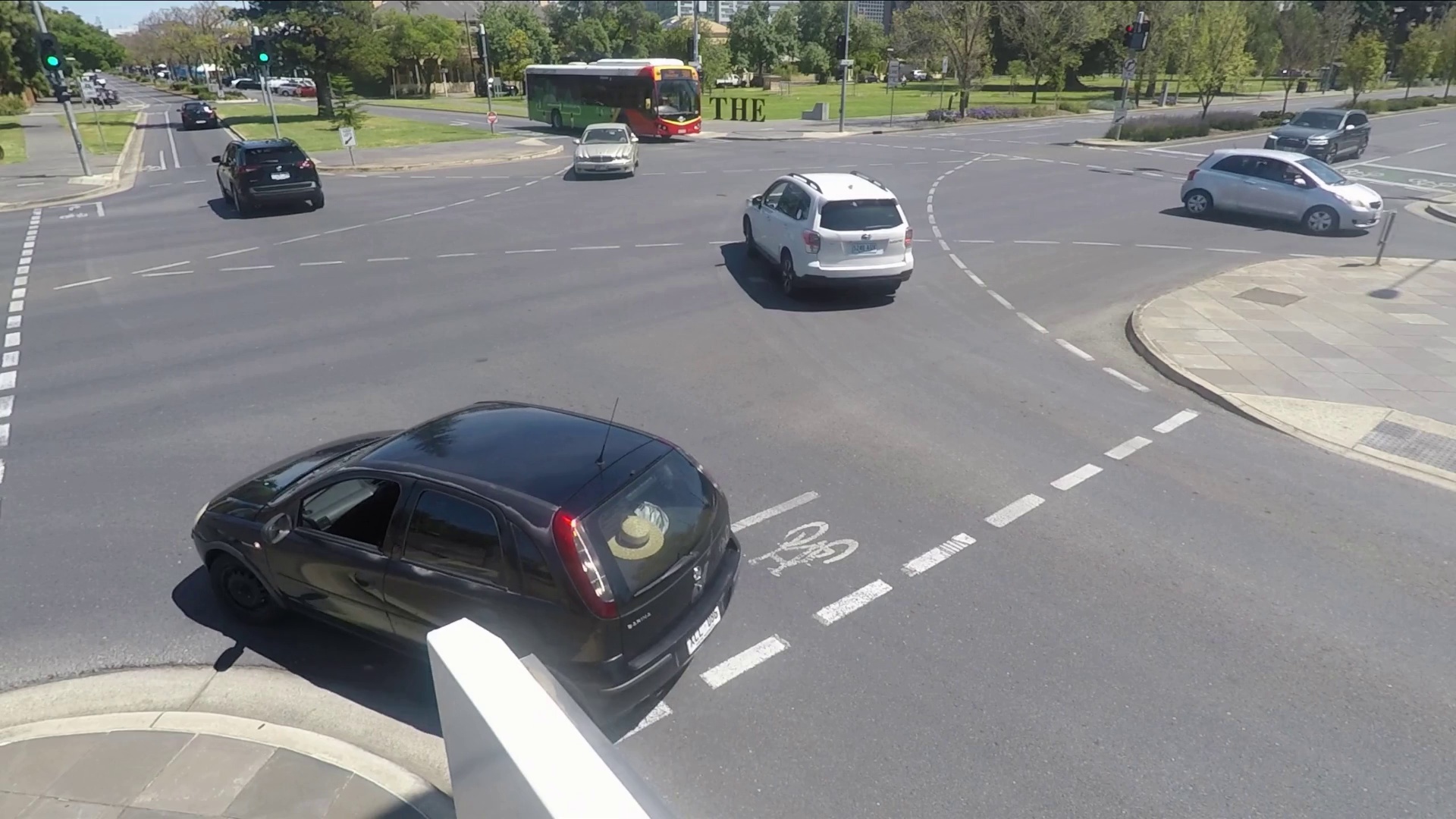}}} \\
        \bmvaHangBox{\fbox{\includegraphics[width=.5\textwidth]{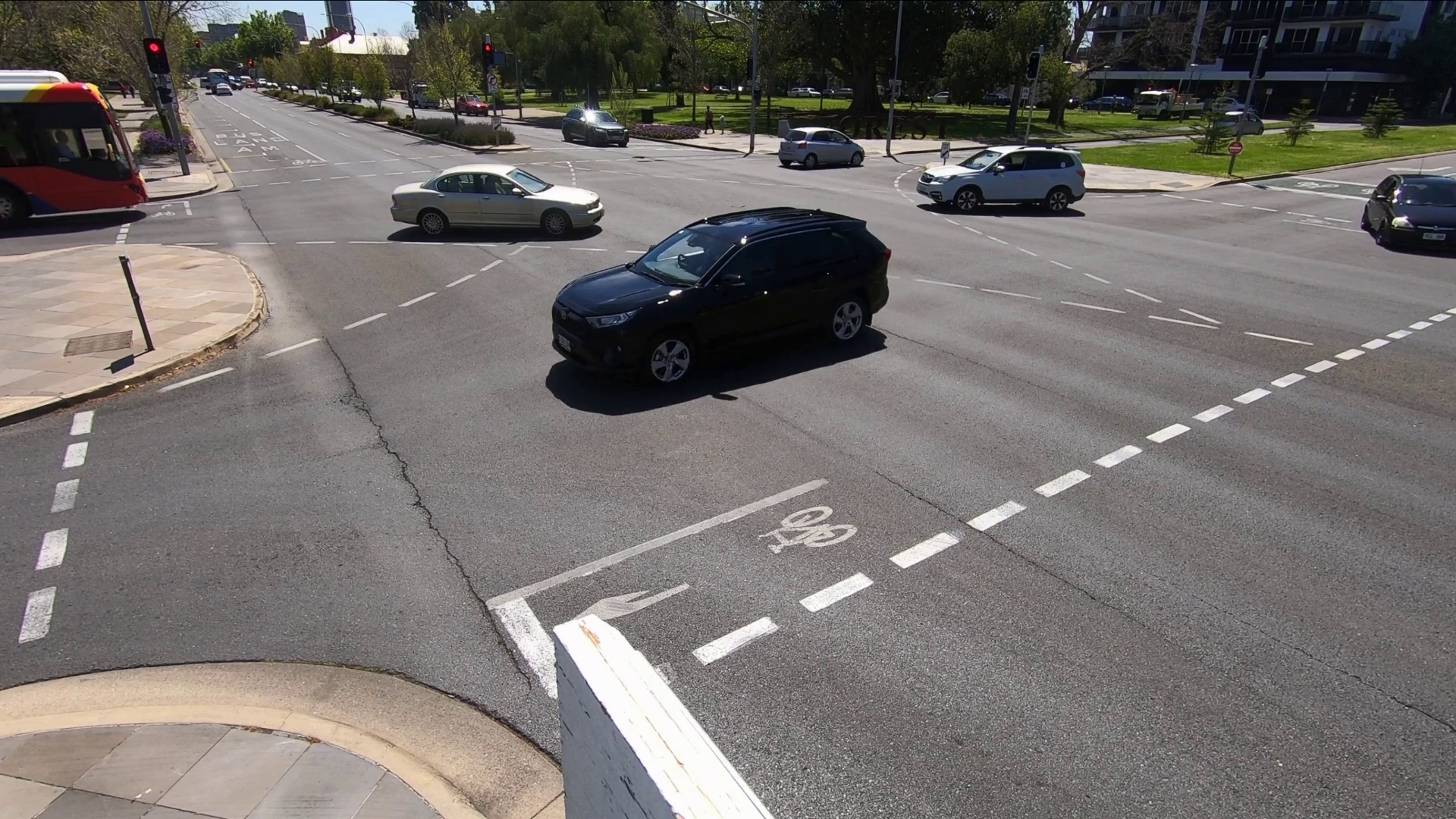}}} &
        \bmvaHangBox{\fbox{\includegraphics[width=.5\textwidth]{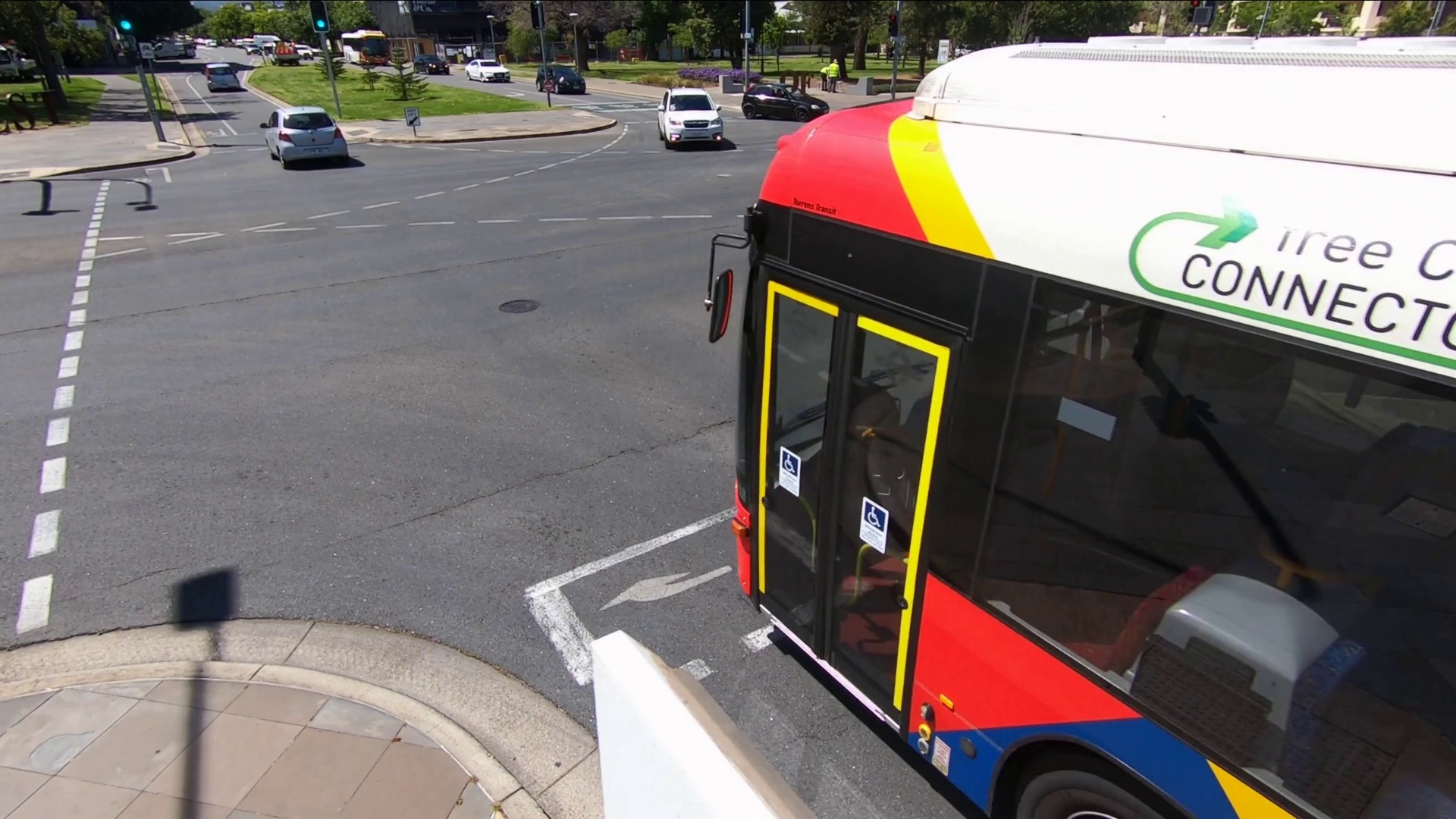}}} \\
        \multicolumn{2}{c}{(a)}\\
        \bmvaHangBox{\fbox{\includegraphics[width=.5\textwidth]{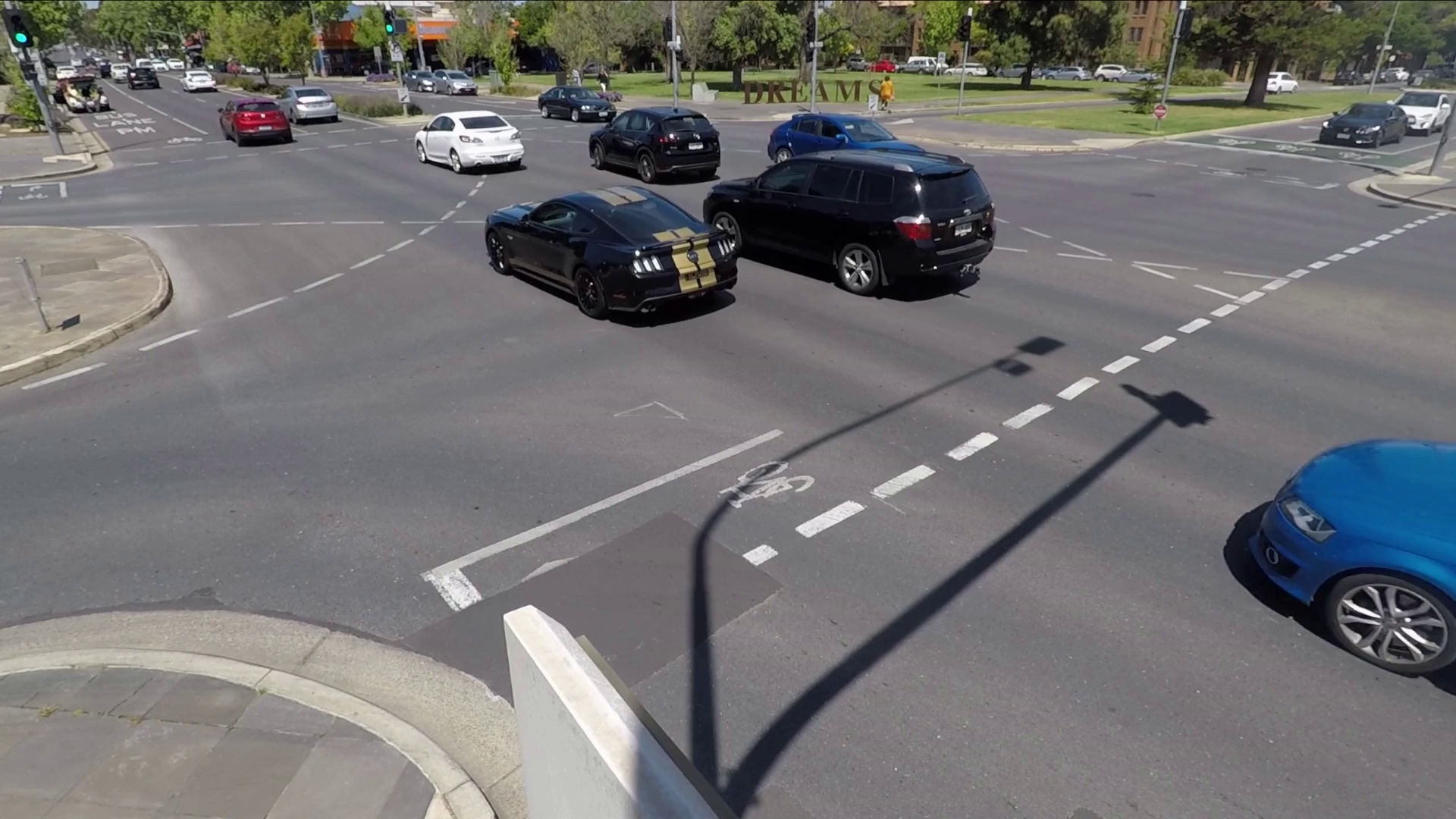}}} &
        \bmvaHangBox{\fbox{\includegraphics[width=.5\textwidth]{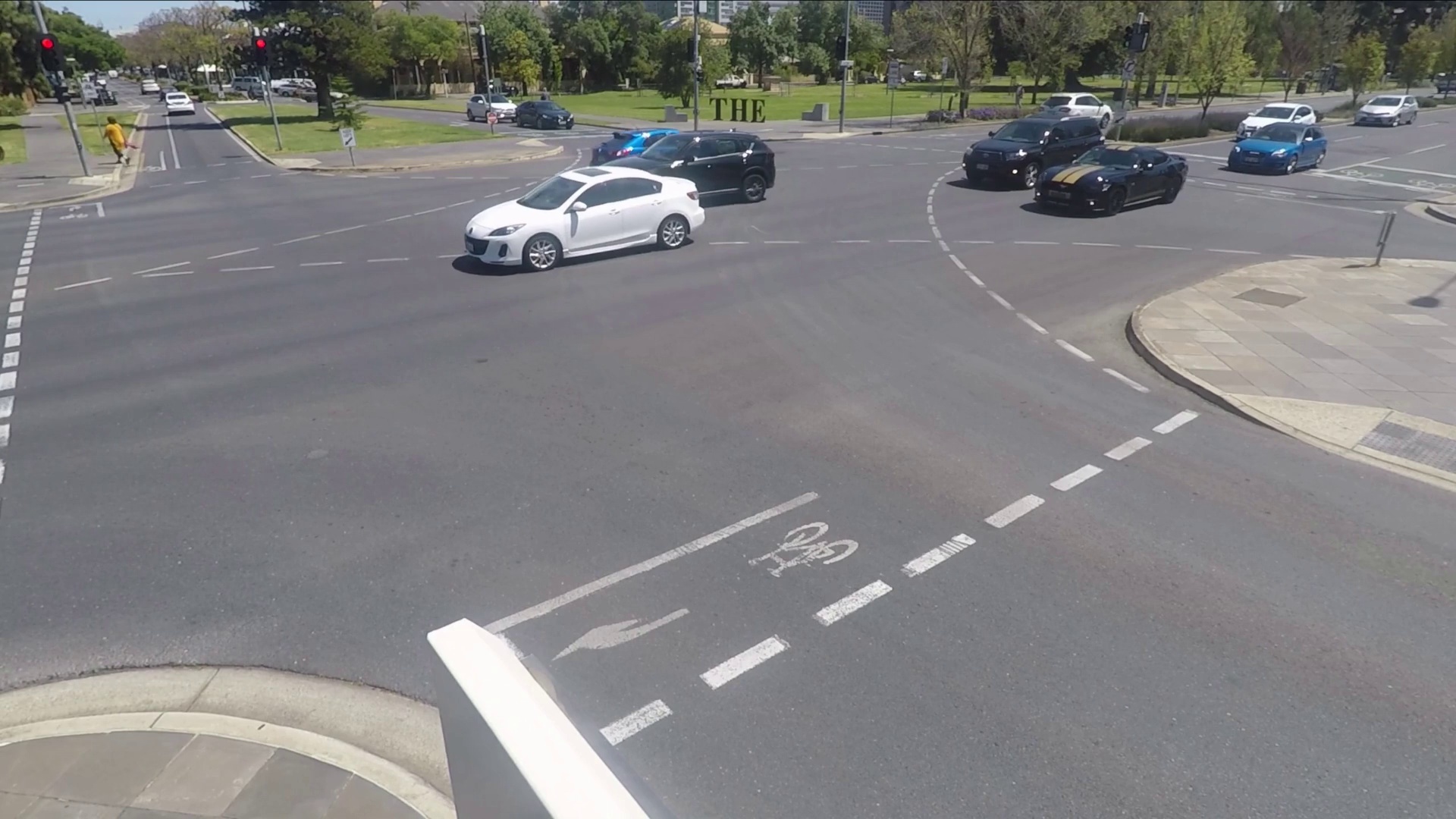}}} \\
        \bmvaHangBox{\fbox{\includegraphics[width=.5\textwidth]{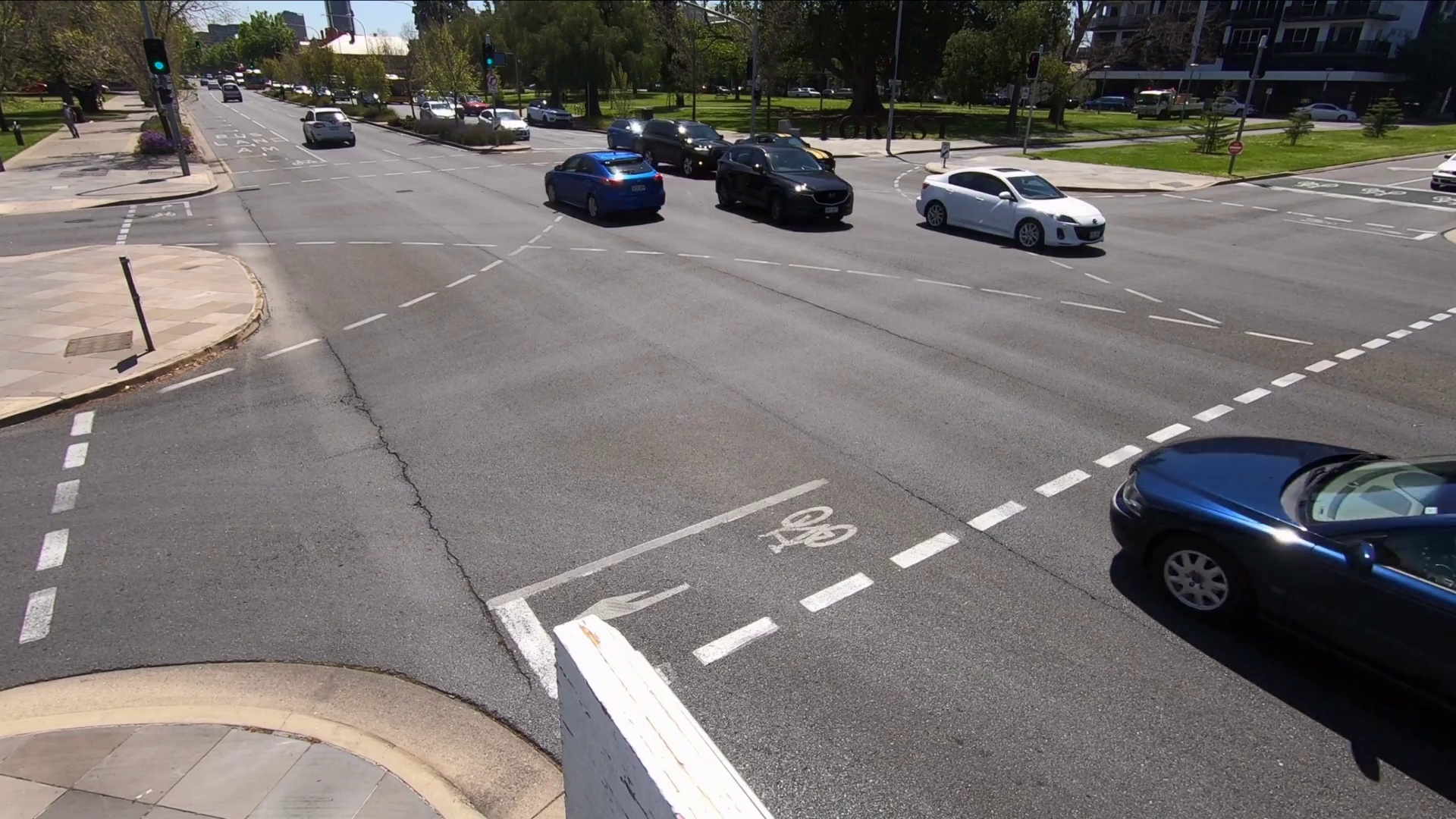}}} &
        \bmvaHangBox{\fbox{\includegraphics[width=.5\textwidth]{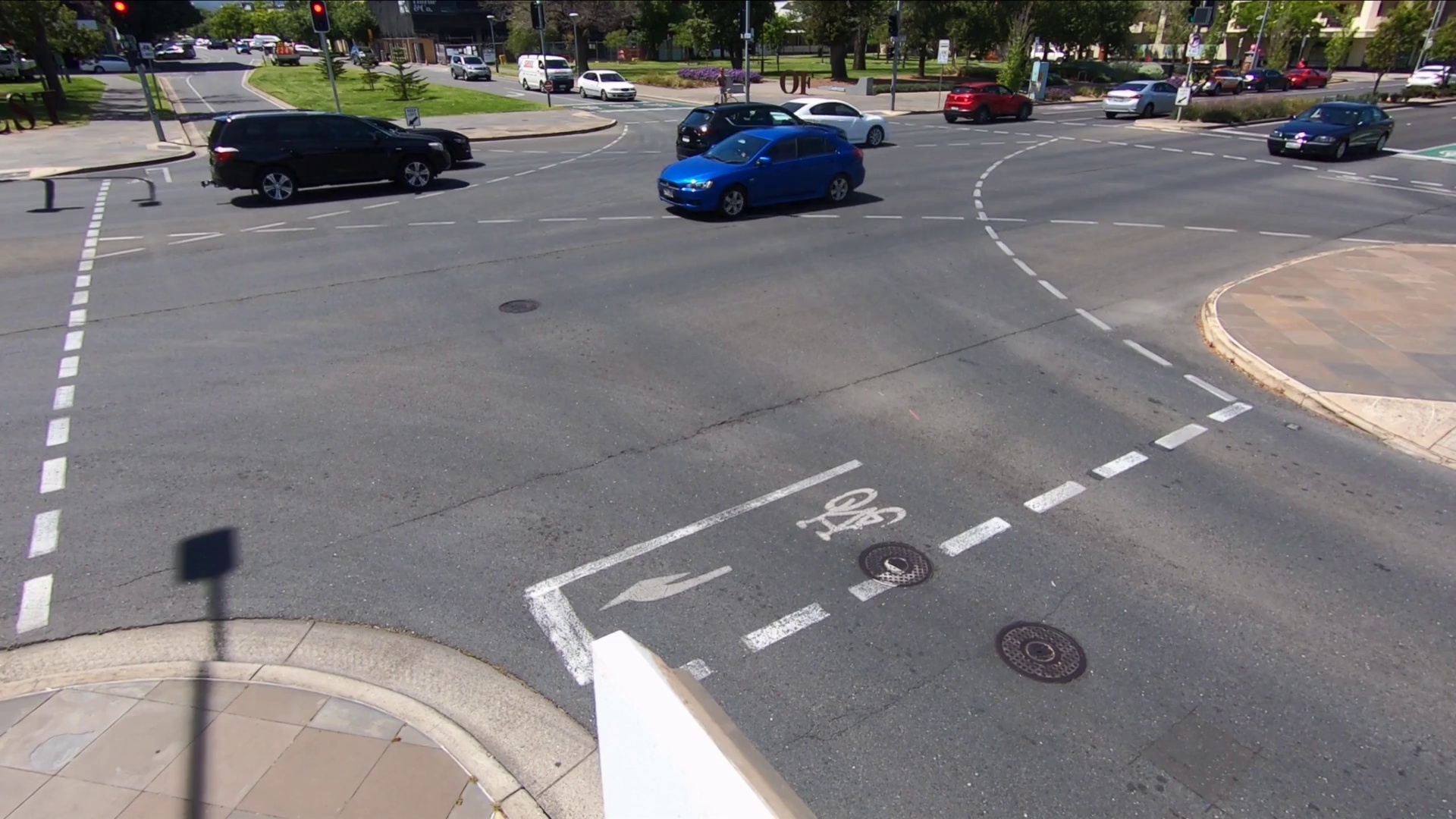}}}\\
        \multicolumn{2}{c}{(b)}
        \end{tabular}
    }
    \caption{Example images of the dataset showing the view from each camera. (a) Sample from the training set (b) Sample from the validation set}
    \label{fig:datasetexamples}
\end{figure}

\begin{figure}[h]
\centering
    \bmvaHangBox{\fbox{\includegraphics[width=0.5\textwidth]{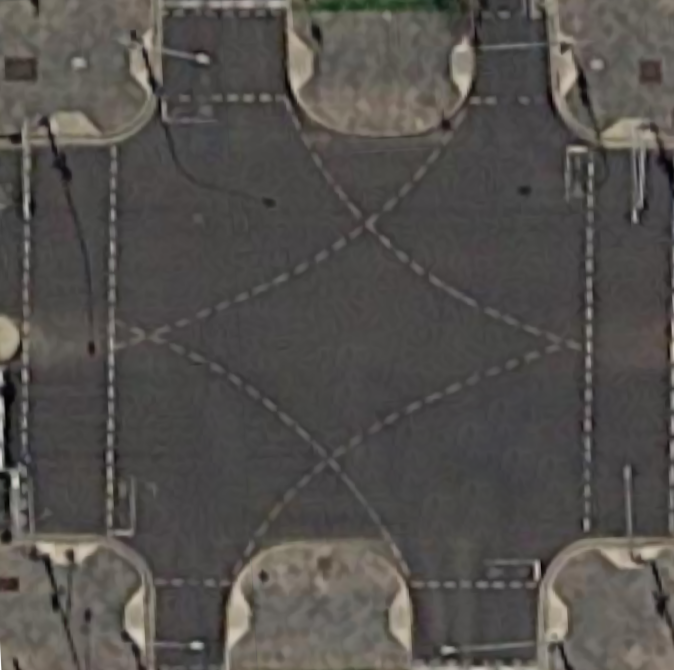}}}
\caption{Satellite image of the intersection being observed. Imagery \copyright2021 Aerometrex Pty Ltd, Map data \copyright2021 Google}
\label{fig:cleanBEV}
\end{figure}

\subsection{Camera synchronisation}\label{supsec:camerasync}
\begin{figure}[H]
\centering
    \bmvaHangBox{\fbox{\includegraphics[width=\textwidth]{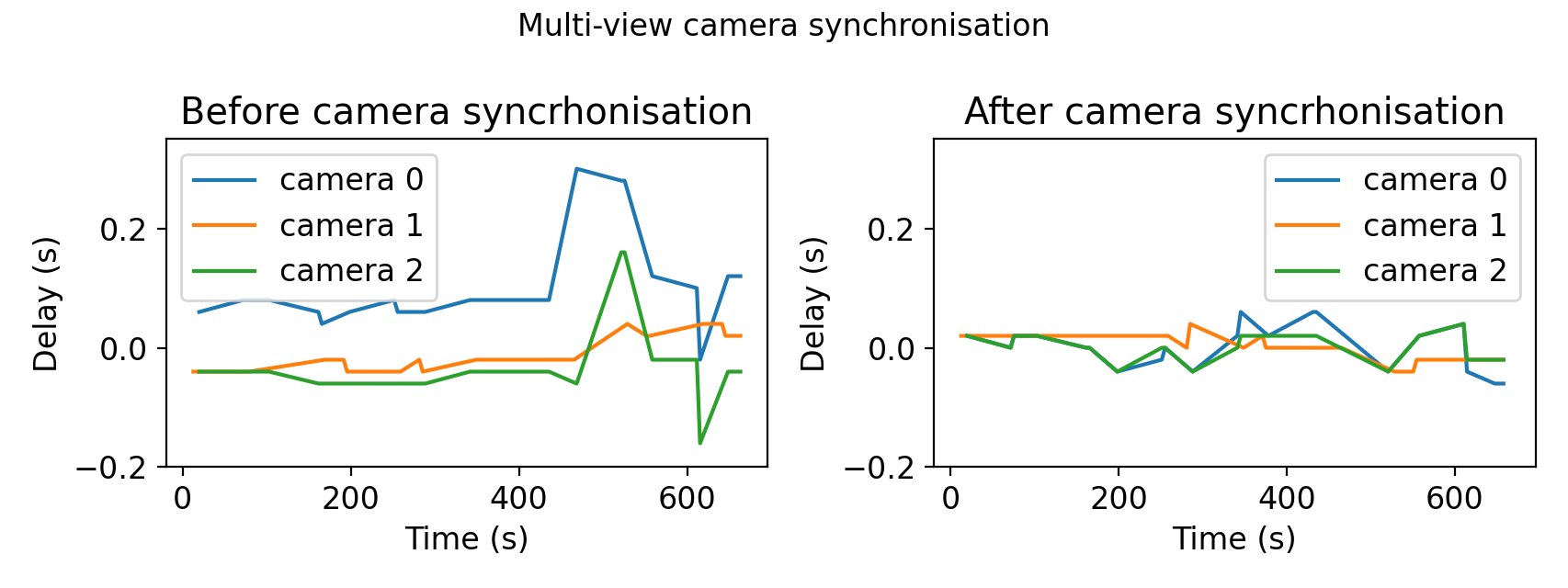}}}
\caption{Before and after synchronising the dataset using the traffic lights changing. Delay with reference to camera 3 in the dataset.}
\label{fig:light_synchronisation}
\end{figure}
Figure \ref{fig:light_synchronisation} shows the improvement in synchronisation between the multi-view cameras using our semi-automatic method. Before synchronisation of cameras, it is clear that there is a large error up to 16 frames towards the end of data collection. This large error can result in cars having zero overlap on the road surface between views. To measure the light changes we can use the change from green to yellow as the measurement signal and yellow to red light change as synchronisation signal. This measures the delay between cameras at the end of the light cycle and leaves a gap between red lights to measure if the cameras go out of sync again.

\subsection{Clustering output}\label{supsec:clustering}
\begin{figure}[H]
    \centering
    \begin{tabular}{c}
    \bmvaHangBox{\includegraphics[width=1\textwidth]{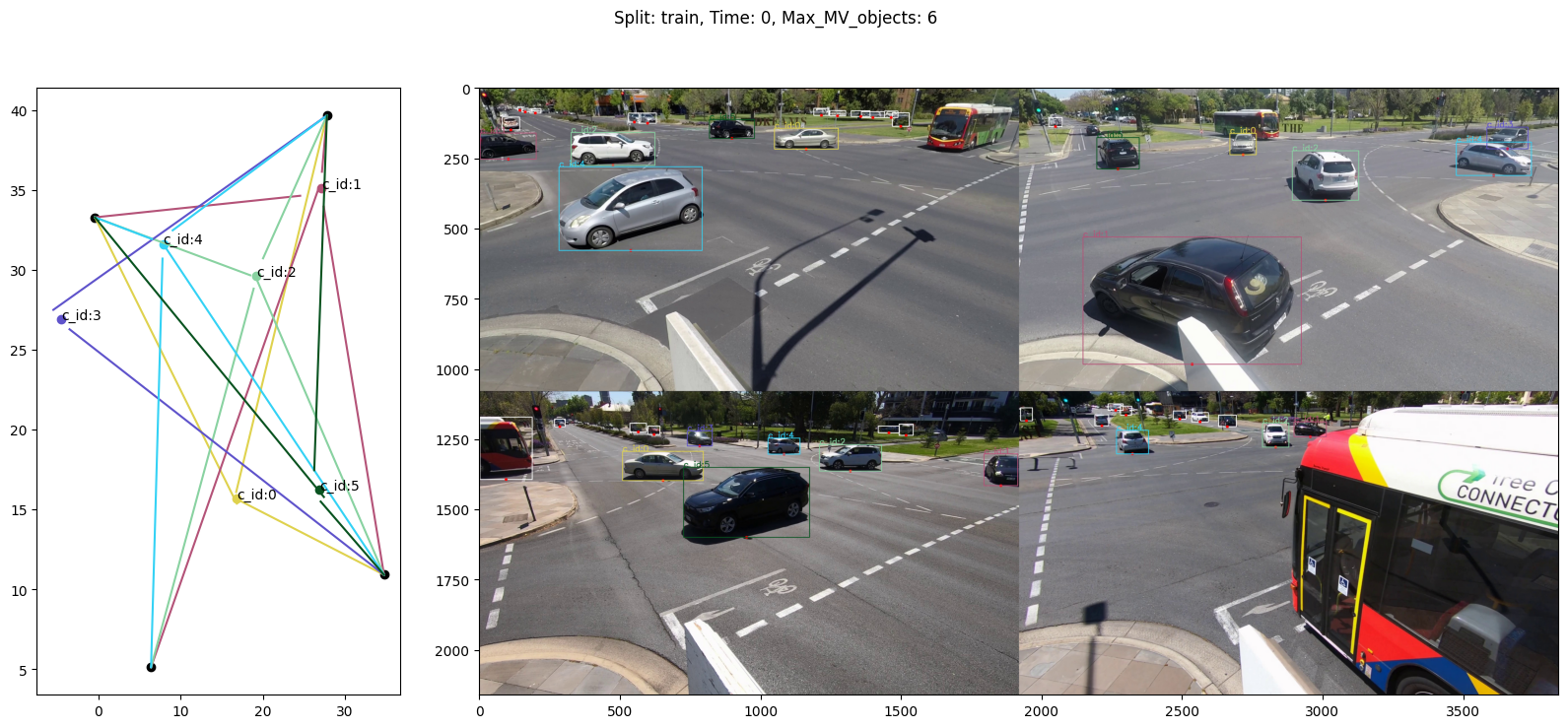}}\\
    (a)\\
    \bmvaHangBox{\includegraphics[width=1\textwidth]{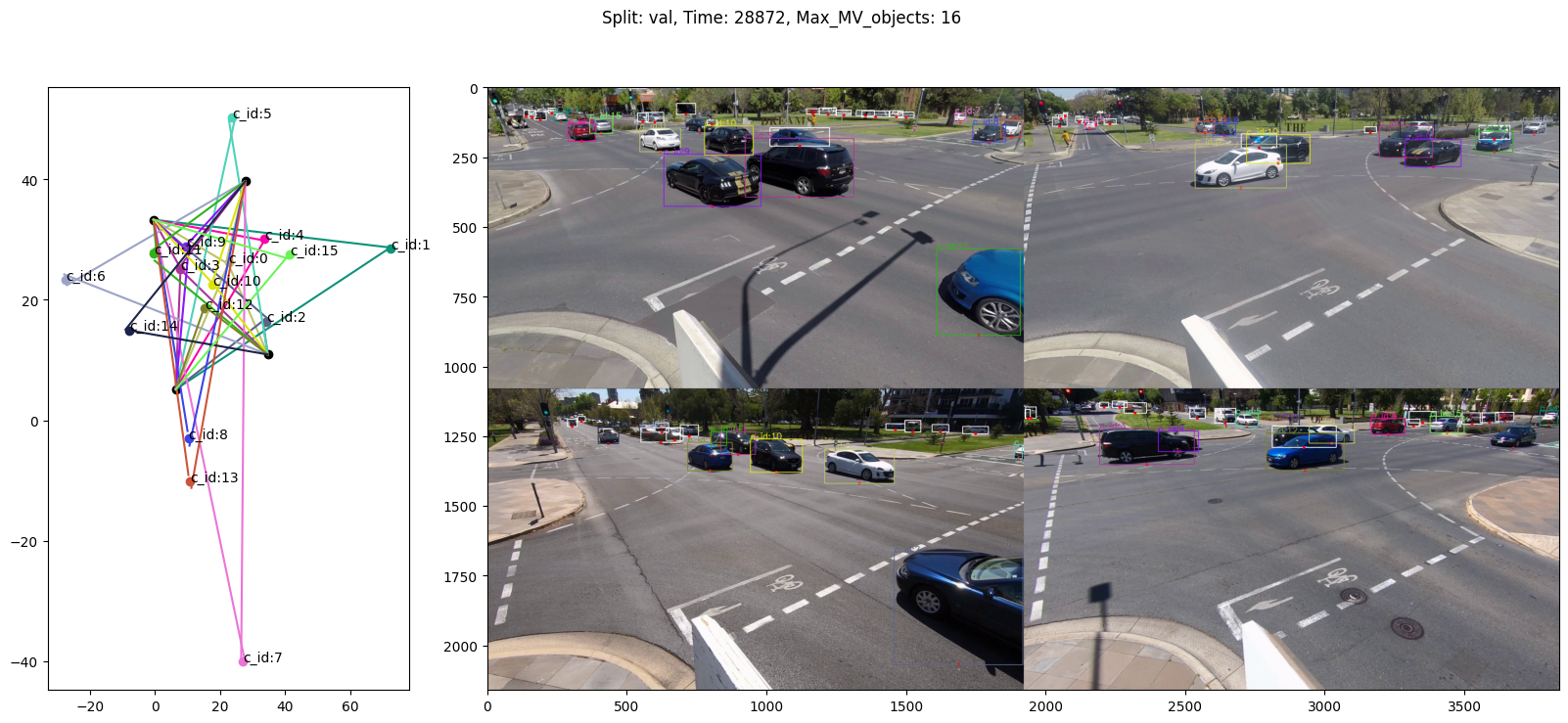}}\\
    (b)
    \end{tabular}
    \caption{(a) Example of the clusters formed with vehicles detected from multiple views (b) Largest number (16) of clusters in the dataset. The colours correspond to a matched cluster. Black dots represent the four cameras with lines drawn out from the camera to the detection centre on the ground plane. Best viewed in colour and zoomed in.}
    \label{fig:clustering examples}
\end{figure}
Figure \ref{fig:clustering examples} shows the output of the clustering algorithm used for the multi-view vehicle associations. The diagram on the left shows the BEV clusters where the black dots are the camera centres with lines to the detection ground points. Objects which are clustered have a unique colour between the detection images on the right and the BEV diagram on the left. Figure \ref{fig:clustering examples}(b) shows the highest number of matched vehicles the system found with 16 unique objects viewed at this time instance.

\subsection{Automatically annotated multi-view vehicle distributions }\label{supsec:WIBAMdistributions}
\begin{figure}[H]
    \centering
    \bmvaHangBox{\fbox{\includegraphics[width=\textwidth]{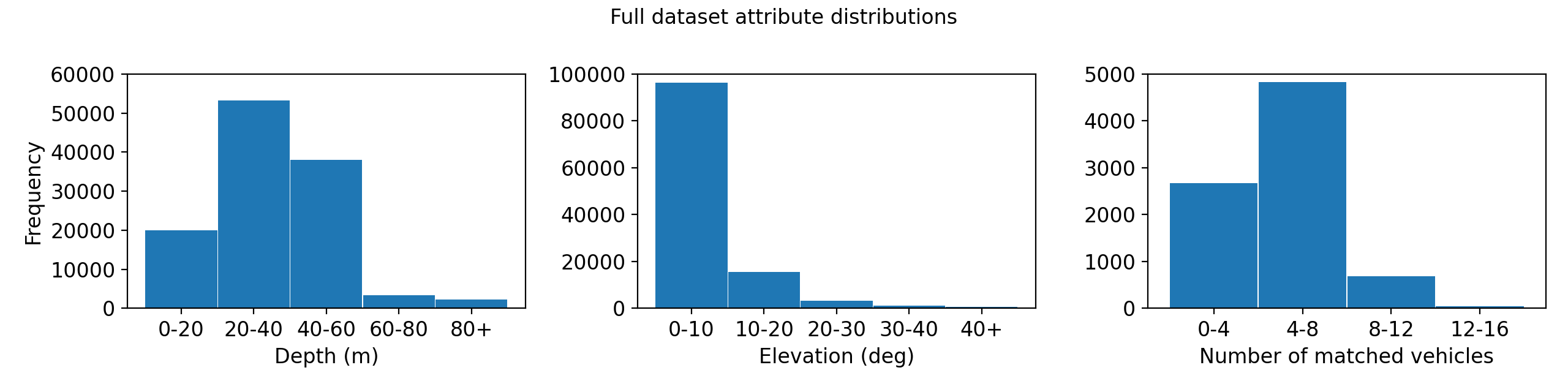}}}
    \caption{Distributions of key data elements within the WIBAM dataset}
    \label{fig:auto_labelled_data_distribution}
\end{figure}

Varying the positions of vehicles within the intersection give a wide range of depths and elevations vehicles are viewed from. There is also a different numbers of vehicles associated and matched across views at any given time, with a maximum of 16 shown in Figure \ref{fig:clustering examples}(b). Figure \ref{fig:auto_labelled_data_distribution} shows the distributions of these in the entire dataset. Our dataset views vehicles at a wide range of elevations, where autonomous vehicle data is typically around 5\textdegree.

\subsection{Test set distributions}\label{supsec:testset}
\begin{figure}[H]
\centering
\bmvaHangBox{\fbox{\includegraphics[width=\textwidth]{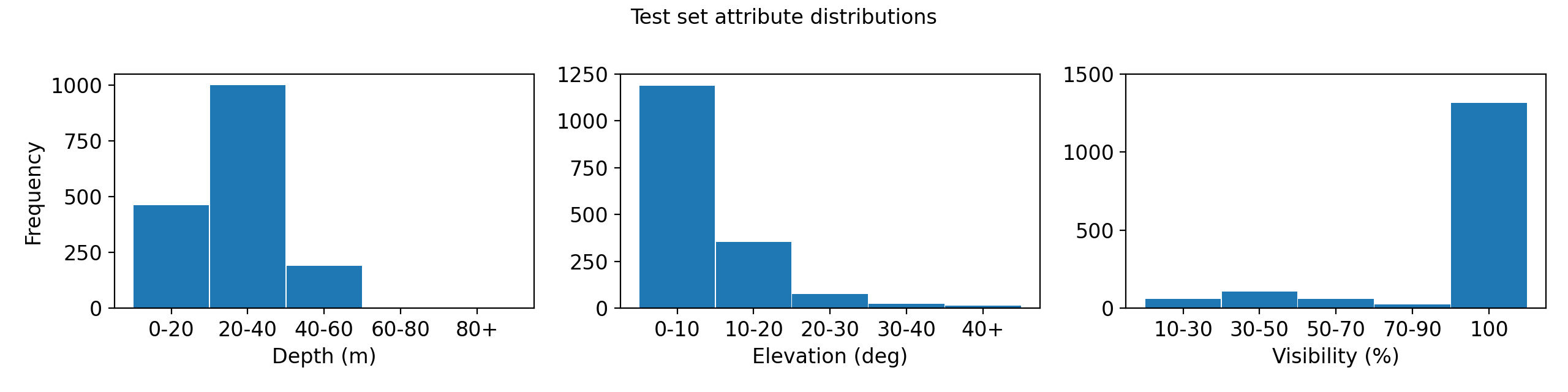}}}
\caption{Hand labelled annotation distributions}
\label{fig:hand-label-distributions}
\end{figure}

Figure \ref{fig:hand-label-distributions} shows the distributions of depth and elevation angle in the hand labelled test set of the WIBAM dataset. The distributions of elevations and distances are representative of the whole dataset however vehicles at larger distances outside of view for 4 cameras are difficult to label accurately hence there is a drop off above 40m+. The high mounted cameras result in less occlusions of vehicles therefore the majority are not occluded at all.

\subsection{Labelling tool GUI}\label{supsec:labellerGUI}
The labelling tool allows the user to annotate vehicles with 7DoF pose. We found that annotating vehicles using this tool takes around 2.5 minutes per time instance or just over half a minute per image labelled. 
\begin{figure}[H]
    \centering
    \begin{tabular}{cc}
        \bmvaHangBox{\fbox{\includegraphics[width=1\textwidth]{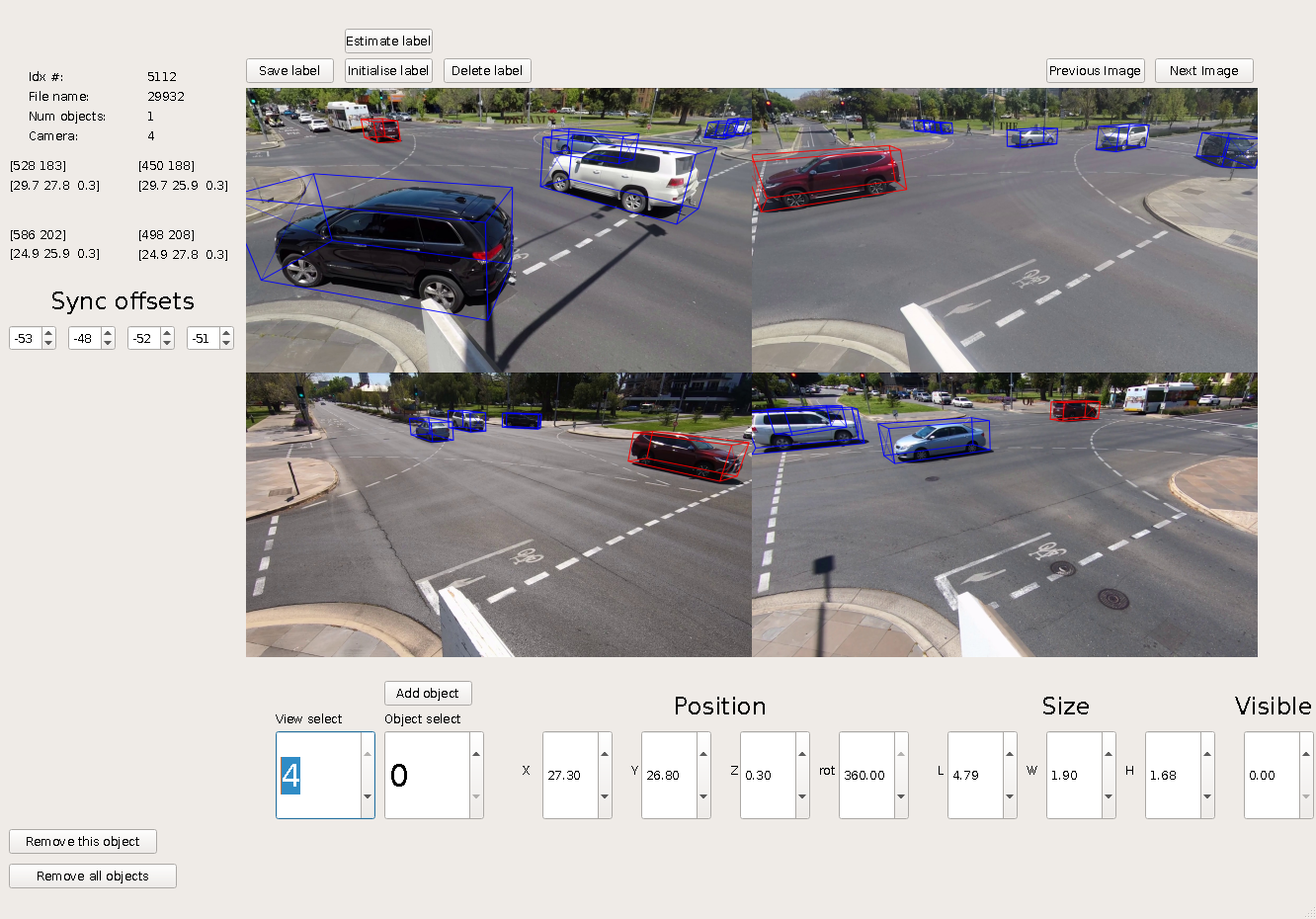}}}&
        \bmvaHangBox{\fbox{\includegraphics[width=1\textwidth]{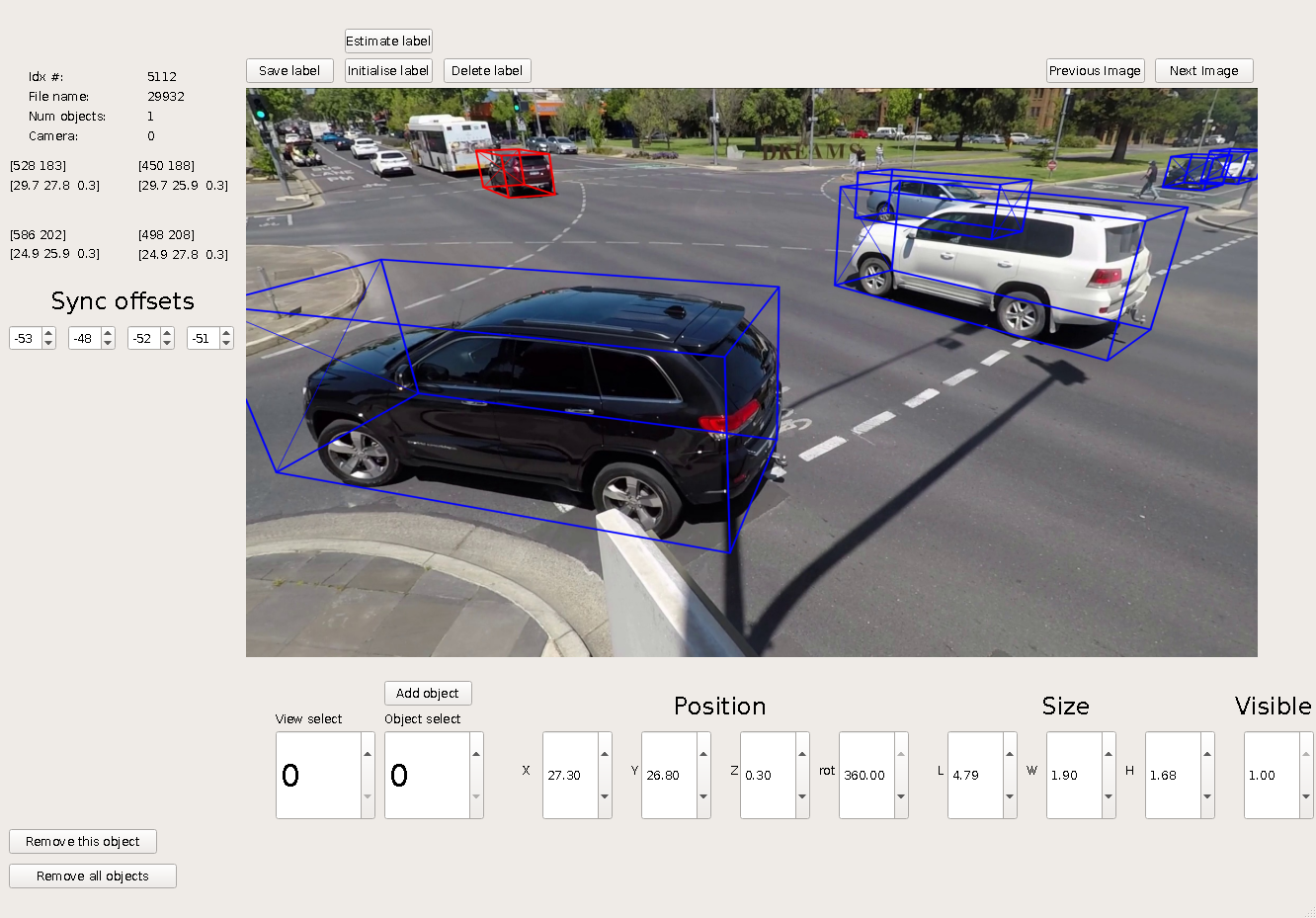}}}\\
        (a)&(b)
    \end{tabular}
    \caption{The labelling tool created for labelling monocular multi-view data}
    \label{fig:labelGUI}
\end{figure}